\pdfoutput=1

\documentclass[journal]{IEEEtran}

\usepackage{cite}
\ifCLASSINFOpdf
  \usepackage[pdftex]{graphicx}
  \graphicspath{{../pdf/}{../jpeg/}}
   \DeclareGraphicsExtensions{.pdf,.jpeg,.png}
\else
   \usepackage[dvips]{graphicx}
   \graphicspath{{../eps/}}
   \DeclareGraphicsExtensions{.eps}
\fi
\ifCLASSOPTIONcompsoc
\usepackage[caption=false,font=normalsize,labelfon
t=sf,textfont=sf]{subfig}
\else
\usepackage[caption=false,font=footnotesize]{subfi
	g}
\fi

\usepackage{amsmath}
\usepackage{amsfonts} 
\usepackage{algorithm}
\usepackage{algorithmic}
\begin{document}

%
% paper title
% Titles are generally capitalized except for words such as a, an, and, as,
% at, but, by, for, in, nor, of, on, or, the, to and up, which are usually
% not capitalized unless they are the first or last word of the title.
% Linebreaks \\ can be used within to get better formatting as desired.
% Do not put math or special symbols in the title.
\title{Boosting in Univariate Nonparametric Maximum Likelihood Estimation}
%
%
% author names and IEEE memberships
% note positions of commas and nonbreaking spaces ( ~ ) LaTeX will not break
% a structure at a ~ so this keeps an author's name from being broken across
% two lines.
% use \thanks{} to gain access to the first footnote area
% a separate \thanks must be used for each paragraph as LaTeX2e's \thanks
% was not built to handle multiple paragraphs
%

\author{YunPeng ~Li\quad ZhaoHui ~Ye
	
	% <-this % stops a space
	\thanks{YunPeng Li was with the Department of Automation, Tsinghua University, Beijing, China e-mail: liyp18@mails.tsinghua.edu.cn}
	\thanks{ZhaoHui Ye was with the Department of Automation, Tsinghua University, Beijing, China e-mail: yezhaohui@mail.tsinghua.edu.cn}
}

% note the % following the last \IEEEmembership and also \thanks - 
% these prevent an unwanted space from occurring between the last author name
% and the end of the author line. i.e., if you had this:
% 
% \author{....lastname \thanks{...} \thanks{...} }
%                     ^------------^------------^----Do not want these spaces!
%
% a space would be appended to the last name and could cause every name on that
% line to be shifted left slightly. This is one of those "LaTeX things". For
% instance, "\textbf{A} \textbf{B}" will typeset as "A B" not "AB". To get
% "AB" then you have to do: "\textbf{A}\textbf{B}"
% \thanks is no different in this regard, so shield the last } of each \thanks
% that ends a line with a % and do not let a space in before the next \thanks.
% Spaces after \IEEEmembership other than the last one are OK (and needed) as
% you are supposed to have spaces between the names. For what it is worth,
% this is a minor point as most people would not even notice if the said evil
% space somehow managed to creep in.

% The paper headers
\markboth{Journal of \LaTeX\ Class Files,~Vol.~14, No.~8, August~2015}%
{Shell \MakeLowercase{\textit{et al.}}: Bare Demo of IEEEtran.cls for IEEE Journals}
% The only time the second header will appear is for the odd numbered pages
% after the title page when using the twoside option.
% 
% *** Note that you probably will NOT want to include the author's ***
% *** name in the headers of peer review papers.                   ***
% You can use \ifCLASSOPTIONpeerreview for conditional compilation here if
% you desire.

% If you want to put a publisher's ID mark on the page you can do it like
% this:
%\IEEEpubid{0000--0000/00\$00.00~\copyright~2015 IEEE}
% Remember, if you use this you must call \IEEEpubidadjcol in the second
% column for its text to clear the IEEEpubid mark.

% use for special paper notices
%\IEEEspecialpapernotice{(Invited Paper)}

% make the title area
\maketitle

% As a general rule, do not put math, special symbols or citations
% in the abstract or keywords.
\begin{abstract}
Nonparametric maximum likelihood estimation is intended to infer the unknown density distribution while making as few assumptions as possible. To alleviate the over parameterization in nonparametric data fitting, smoothing assumptions are usually merged into the estimation. In this paper a novel boosting-based method is introduced to the nonparametric estimation in univariate cases. We deduce the boosting algorithm by the second-order approximation of nonparametric log-likelihood. Gaussian kernel and smooth spline are chosen as weak learners in boosting to satisfy the smoothing assumptions. Simulations and real data experiments demonstrate the efficacy of the proposed approach.     
\end{abstract}

% Note that keywords are not normally used for peerreview papers.

% For peer review papers, you can put extra information on the cover
% page as needed:
% \ifCLASSOPTIONpeerreview
% \begin{center} \bfseries EDICS Category: 3-BBND \end{center}
% \fi
%
% For peerreview papers, this IEEEtran command inserts a page break and
% creates the second title. It will be ignored for other modes.
\IEEEpeerreviewmaketitle

\section{Introduction}
% The very first letter is a 2 line initial drop letter followed
% by the rest of the first word in caps.
% 
% form to use if the first word consists of a single letter:
% \IEEEPARstart{A}{demo} file is ....
% 
% form to use if you need the single drop letter followed by
% normal text (unknown if ever used by the IEEE):
% \IEEEPARstart{A}{}demo file is ....
% 
% Some journals put the first two words in caps:
% \IEEEPARstart{T}{his demo} file is ....
% 
% Here we have the typical use of a "T" for an initial drop letter
% and "HIS" in caps to complete the first word.
\IEEEPARstart{N}{onparametric} maximum likelihood estimation (NPMLE)\cite{jkjw1956,steven1980,alan1991} has received much attention in recent years. It has been successfully applied to various problems in signal processing, statistical learning, and pattern recognition. Given finite independent identically distributed  random samples from an unknown distribution, the goal of NPMLE is to estimate the probability density with as few assumptions as possible. Unfortunately, NPMLE's optimization over an infinite-dimensional function space often leads to the unbounded likelihood and overfitting. The remedy to alleviate these defects is merging additional assumptions or constraints into the estimation. These assumptions confine the nonparametric density estimation to certain functional spaces. One of the most popular methods is to impose the smoothing constraint on the unknown distribution to restrict the estimation.
\par There are currently two common approaches to utilize the smoothing constraint: restriction methods and
regularization methods. Conventional restriction methods control the smoothing degree via predetermined smoothing parameters (such as the number of bins in the histogram, the number of observations in the nearest-neighbor method, the bandwidth in kernel methods\cite{eman1962,nada1965,jeff1993} and the local polynomials\cite{Loader1999,matia2020}). Another kind of restriction methods supposes the unknown distribution owns special structures like mixture models\cite{nan1978,redner1984,mich1995,pri2018},
shape constraints (log-concavity\cite{samworth2018} and monotonicity\cite{minyen1999}). In regularization methods, penalty terms (such as roughness\cite{goodd1971,bws1982}, $L_{1}$ penalty\cite{bunea2007}, total variation\cite{koenker2006} ) are designed to control the smoothing degree. For roughness penalty, nonparametric maximum penalty likelihood is analyzed in the reproducing kernel Hilbert spaces\cite{demontricher1975} and one of its estimates is proven to be a positive exponential smooth spline\cite{bookSpline} with knots only at the sample points\cite{demontricher1975,Edward1983}.
\par However, most of the restriction and selection methods have to determine the tuning parameters beforehand\cite{jeff1993,Gu03}, resulting in a lack of flexibility in inference. In this paper, a selection method is introduced to NPMLE in the boosting form. The proposed algorithm adaptively scans the function spaces and includes only those that contribute significantly to estimation. 

\par Our contributions are as follows. 
\begin{enumerate}
\item We derive the boosting algorithm from the second-order approximation of nonparametric log-likelihood. 
\item We select several weak learners for boosting NPMLE. Different from the regularization methods, those weak learners share the fixed smoothing degree at each iteration. The only meta-parameter in boosting NPMLE is the number of boosting iterations. 
\end{enumerate}

\section{Proposed method}
\par Let $X$ be a random variable in $\mathbb{R}$ with probability density $p(x)$. Given $N$ independent identically distributed samples $X_{1},X_{2},\cdots,X_{N}$, we model the density estimate $\hat{p}(x)$ in the form of Gibbs distribution.
\begin{equation}
\label{gibbs}
\mit{\hat{p}(x)}=\frac{e^{f(x)}}{\int e^{f(x)}dx}
\end{equation}
where $f(x)$ is assumed to be a  smooth function in $\mathbb{R}$. The log likelihood $L(f)$ is defined as the function of $f(x)$.
\begin{equation}
\label{loglik1}
L(f) =\frac{1}{N}\sum_{i=1}^{N}log\,\hat{p}(X_{i})
\end{equation}
Supposing that $x_{1},x_{2},\cdots,x_{n}$ are the $n$ unique elements of $X_{1},X_{2},\cdots,X_{N}$ in ascending order, we calibrate their frequencies $q_{i}$ as
\begin{equation}
\label{frqs}
q_{i}=\frac{1}{N} \#\{j\leq N|X_{j}=x_{i}\}
\end{equation}
  where $\#$ is the counting function. We restrict the support of $\hat{p}(x)$ in $[x_{1},x_{n}]$. The trapezoidal rule is used for numerical integration in equation (\ref{gibbs}), where $a_{i}$ is the coefficient concerning $x_{i}$. Then, the estimation $\hat{p}(x)$ and the log-likelihood $L(f)$ are adjusted in the following form
\begin{equation}
\label{gibbs2}
\hat{p}(x)=\frac{e^{f(x)}}{\sum_{i=1}^{n} a_{i}e^{f(x_{i})}}
\end{equation}
\begin{equation}
\label{loglik12}
\begin{aligned}
L(f) &=\sum_{i=1}^{n}q_{i}log\,\hat{p}(x_{i})\\
&=\sum_{i=1}^{n}q_{i}f(x_{i})-log\, \sum_{i=1}^{n}a_{i}e^{f(x_{i})}
\end{aligned}
\end{equation}

\par To avoid the summation in logarithm in equation (\ref{loglik12}), we replace the original $L(f)$ with a simpler
surrogate $\mathcal{L}(f)$ according to the inequality $log\,v\leq -1+v$.
\begin{equation}
\label{loglik4}
\mathcal{L}(f) =\sum_{i=1}^{n}q_{i}f(x_{i})- \sum_{i=1}^{n}a_{i}e^{f(x_{i})}
\end{equation}
\begin{equation}
\label{loglik3}
\begin{aligned}
L(f) &\ge 1+\sum_{i=1}^{n}q_{i}f(x_{i})- \sum_{i=1}^{n}a_{i}e^{f(x_{i})}\\
&\ge 1+\mathcal{L}(f) 
\end{aligned}
\end{equation}
It can be proved that the original $L(f)$ and surrogate $\mathcal{L}(f)$ have an identical maximum point. Thus, we optimize the surrogate $\mathcal{L}(f)$ as the objective function in the NPMLE.
\par In the remaining part, We firstly derive the boosting algorithm to optimize $\mathcal{L}(f)$. Then we select several weak learners to validate the proposed method.
%\begin{equation}
%\label{gaps}
%\mit{a_{i}}=
%\begin{cases}
%\frac{1}{2}(\mit{x_{2}}-\mit{x_{1}})& \text{i=1}\\
%\frac{1}{2}(\mit{x_{i+1}}-\mit{x_{i-1}})& \text{i=2,$\cdots$,n-1}\\
%\frac{1}{2}(\mit{x_{n}}-\mit{x_{n-1}})& \text{i=n}
%\end{cases}
%\end{equation}
\subsection{Boosting  NPMLE}
\par Boosting\cite{boostingref} is a technique of combining multiple weak learners to produce a powerful committee, whose performance is significantly better than any of the weak learners. It works by applying the weak learner sequentially to a dataset of weighted form. For applying the boosting principle to NPMLE, we express $f(x)$ as a combination of weak learner $b(x;\gamma_{m})$
\begin{equation}
\label{boosting1}
f(x)=\sum_{m=1}^{M}b(x;\gamma_{m})
\end{equation}
where $M$ is the number of boosting iterations and $m$ is the index of the single iteration. At each iteration, we train a single weak learner $b(x;\gamma_{m})$ on the weighted data, characterized by a set of parameters $\gamma_{m}$. Thus, the maximum log-likelihood in $\mathcal{L}(f)$ is changed into a boosting form
\begin{equation}
\max_{\{\gamma_{m}\}_{1}^{M}}\quad \mathcal{L}(\sum_{m=1}^{M}b(x;\gamma_{m}))
\end{equation}
\par We define the $f(x)$ at $m$ iteration as $f_{m}(x)$.
\begin{equation}
\label{boosting3}
\begin{aligned}
f_{m}(x)&=\sum_{i=1}^{m}b(x;\gamma_{i})\\
&=f_{m-1}(x)+b(x;\gamma_{m})
\end{aligned}
\end{equation}
The key of boosting is that no earlier parameters $\gamma$ are adjusted at the current $m$ iteration. To acquire $f_{m}(x)$, we optimize a subproblem based on former $f_{m-1}(x)$ sequentially. 
\begin{equation}
\label{boosting2}
\max_{\gamma_{m}}\quad \mathcal{L}(f_{m-1}(x)+b(x;\gamma_{m}))
\end{equation}

A second-order approximation $\mathcal{L}(f_{m};f_{m-1})$ is used to solve  $\mathcal{L}(f_{m})$ on equation (\ref{boosting2}),
\begin{equation}
\begin{aligned}
\label{boosting4}
\mathcal{L}(f_{m};f_{m-1})&\approx \mathcal{L}(f_{m-1}(x)) \\
+&\sum_{i=1}^{n}(q_{i}-a_{i}e^{f_{m-1}(x_{i})})(f_{m}(x_{i})-f_{m-1}(x_{i}))\\
+&\sum_{i=1}^{n}\frac{1}{2}(-a_{i}e^{f_{m-1}(x_{i})})(f_{m}(x_{i})-f_{m-1}(x_{i}))^{2}
\end{aligned}
\end{equation}
Maximizing $\mathcal{L}(f_{m};f_{m-1})$ is equivalent to the minimizing of weighted least squares problem as follow,
\begin{equation}
\begin{aligned}
\label{wlse}
\min_{\gamma_{m}}\quad \sum_{i=1}^{n}\frac{1}{2}\omega_{m}^{i}(b(x_{i};\gamma_{m})-g_{m}(x_{i}))^{2}
\end{aligned}
\end{equation}
\begin{equation}
\label{weight}
\omega_{m}^{i} =a_{i}e^{f_{m-1}(x_{i})}
\end{equation}
\begin{equation}
\label{response}
g_{m}(x_{i}) =\frac{q_{i}-\omega_{m}^{i}}{\omega_{m}^{i}}
\end{equation}
where $\omega_{m}^{i}$ is the weight and $g_{m}(x_{i})$ is the response of ${x_{i}}$ in the current $m$ iteration. For the next $m+1$ iteration, the updating rules concerning the weight and response are
\begin{equation}
\label{response2}
\omega_{m+1}^{i} =\omega_{m}^{i}e^{b(x_{i};\gamma_{m})}
\end{equation}
\begin{equation}
\label{response2}
g_{m+1}(x_{i}) =\frac{q_{i}-\omega_{m+1}^{i}}{\omega_{m+1}^{i}}
\end{equation}
\begin{figure}[!t]
	\centering
	\includegraphics[width=3.5in]{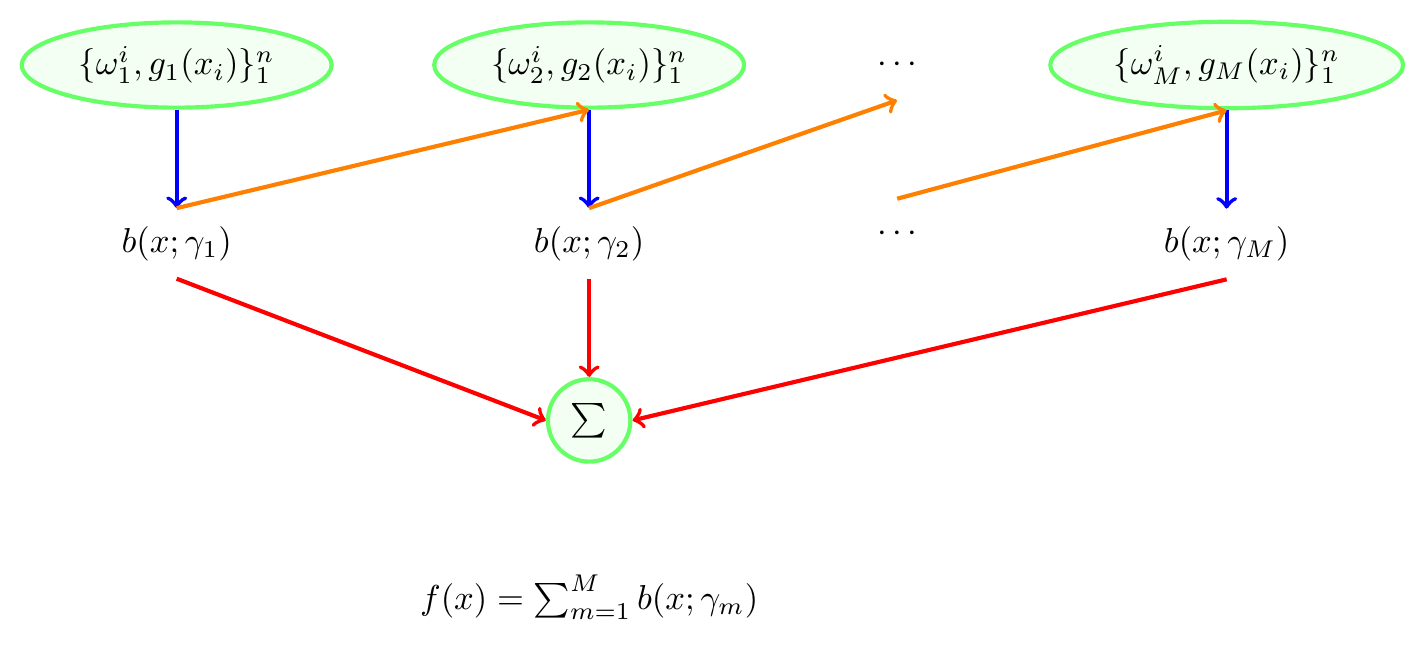}
	\caption{Schematic of boosting in NPMLE. Weak learners are trained on the updated data depend on the performance of the previous iterations, and then combined to produce the final model.}
	\label{boostingchart}
\end{figure}
The parameters $\gamma_{m}$ in single $b(x;\gamma_{m})$ are determined by equation (\ref{wlse}). Once all the weak learners $b(x;\gamma)$ have been trained, $f(x)$ is the combination of whole $M$ weak learners, as illustrated schematically in Fig. \ref{boostingchart}.
The whole algorithm is summarized in Alg. \ref{alg:boosting}. Different from existing boosting algorithm in classification\cite{FREUND1997119} and regression \cite{friedman2001}, boosting NPMLE updates the weight and response of data simultaneously.

\begin{algorithm}[!t]  
	\caption{boosting NPMLE}  
	\label{alg:boosting}  
	\begin{algorithmic}[1]
		\STATE{Initialization}  
		\STATE {$\omega_{0}^{i}\gets \frac{1}{n}$}
		\STATE {$b(x_{i};\gamma_{0})\gets 0$}
		\STATE {$f_{0}(x_{i})\gets 0$}
        \FOR {$m = 1$ to $M$}          
        \FOR {$i = 1$ to $n$}      
        \STATE {$\omega_{m}^{i} \gets\omega_{m-1}^{i}e^{b(x_{i};\gamma_{m-1})}$}
         \STATE {$g_{m}(x_{i}) \gets\frac{q_{i}-\omega_{m}^{i}}{\omega_{m}^{i}}$}
        \ENDFOR
         \STATE{compute}
         \STATE {$\min_{\gamma_{m}}\, \sum_{i=1}^{n}\frac{1}{2}\omega_{m}^{i}(b(x_{i};\gamma_{m})-g_{m}(x_{i}))^{2}$}
        \STATE{$f_{m}(x)\gets f_{m-1}(x)+b(x;\gamma_{m})$}
        \ENDFOR
        \STATE{output $f(x)\gets f_{M}(x)$}
	\end{algorithmic}  
\end{algorithm}  

% needed in second column of first page if using \IEEEpubid
%\IEEEpubidadjcol

\subsection{Choices of weak learners}
\par The choices of the weak learners $b(x;\gamma_{m})$ and the number of boosting  iterations $M$ are the key to boosting NPMLE. Although conventional classification and regression trees(CART)\cite{FREUND1997119,friedman2001,boostingref} can solve equation (\ref{wlse}) efficiently, CART cannot satisfy the smoothing constraint required in boosting NPMLE due to its feature of piecewise constant. Despite boosting method usually reduces training error as the increase of boosting iterations $M$, it can sometimes cause overfitting on future predictions.
\par In boosting NPMLE, ideal weak learners should meet several requirements:
\begin{enumerate}
	\item the weak learners should satisfy the smoothing constraint in NPMLE to avoid over parameterization.
	\item the weak learners can solve the weighted least squares in equation (\ref{wlse}) efficiently.
	\item the model complexity of the weak learners can be easily restricted during each boosting iteration $m$. 
	\item the weak learners should be robust to the large choice of boosting iterations $M$.
\end{enumerate}
\par We select the Gaussian kernel and the smooth spline as weak learners in boosting NPMLE for the following reasons: 
\begin{enumerate}
	\item Gaussian kernel: kernel functions are basis functions for nonlinear mapping, determined by kernel choice and bandwidth. An $L_{2}$ penalty is added to equation (\ref{wlse}) to control their model complexity by the lagrange multiplier $lambda$. These models change from the simple fit to ordinary least squares as the decrease of $lambda$. We select the extremely large choice of $lambda$ ($lambda=10^4$) in boosting NPMLE to constrain their model complexity\cite{glmnet2011}. 
	\item smooth spline: different from CART method, smooth spline is piecewise cubic polynomials under the smoothing constraint. It has been applied and analyzed in nonparametric estimation in regularization methods\cite{demontricher1975,Edward1983}. We fix the complexity of smooth spline via a parameter named degree of freedom
	$df$ ($df=3$). With the increase of $df$ from $2$ to $n$, the $b(x;\gamma_{m})$ changes from a simple line fit to ordinary least squares interpolation.
\end{enumerate}
\par The selection methods in the proposed paper do not focus on the smoothing parameters for single $b(x;\gamma_{m})$. We only determine these weak learners to be extremely simple in each iteration by the fixed $lambda$ (Gaussian kernel) or $df$ (smooth spline). This is the fundamental difference between the existing regularization methods\cite{goodd1971,bws1982,bunea2007,koenker2006}. As a result, our algorithm avoids the difficult choices of tuning parameters. Besides, the extreme choices of large $lambda$ and small $df$ enable boosting NPMLE robust to the large boosting iterations $M$.

\section{Experiments and results}
\begin{table}[]
	\caption{Implementation details}
	\label{details}
	\centering
	\begin{tabular}{l|l|l}
		\hline
		weak learners & package        & parameters                                                                                                                   \\ \hline
		CART          & rpart          & $minsplit$ = 30.                                                                                                                 \\ \hline
		Gaussian kernel        & density;glmnet & \begin{tabular}[c]{@{}l@{}}$kernel$ ="gaussian";$alpha$ = 0,\\ $family$ = "gaussian",$lambda$ = $10^{4}$.\end{tabular} \\ \hline
		smooth spline & smooth.spline  & $df$ = 3.                                                                                                                      \\ \hline
	\end{tabular}
\end{table}
In this section, simulations, and experiment on real data are designed to verify the performance of boosting NPMLE in univariate cases. The only tuning parameter is the number of boosting iterations $M$, more details are shown in Table \ref{details}.
\subsection{Improvement in data fitting}
\par
\begin{figure*}[!t]
	\centering
	\subfloat[Boosting iteration $M = 1$]{\includegraphics[width=3.5in]{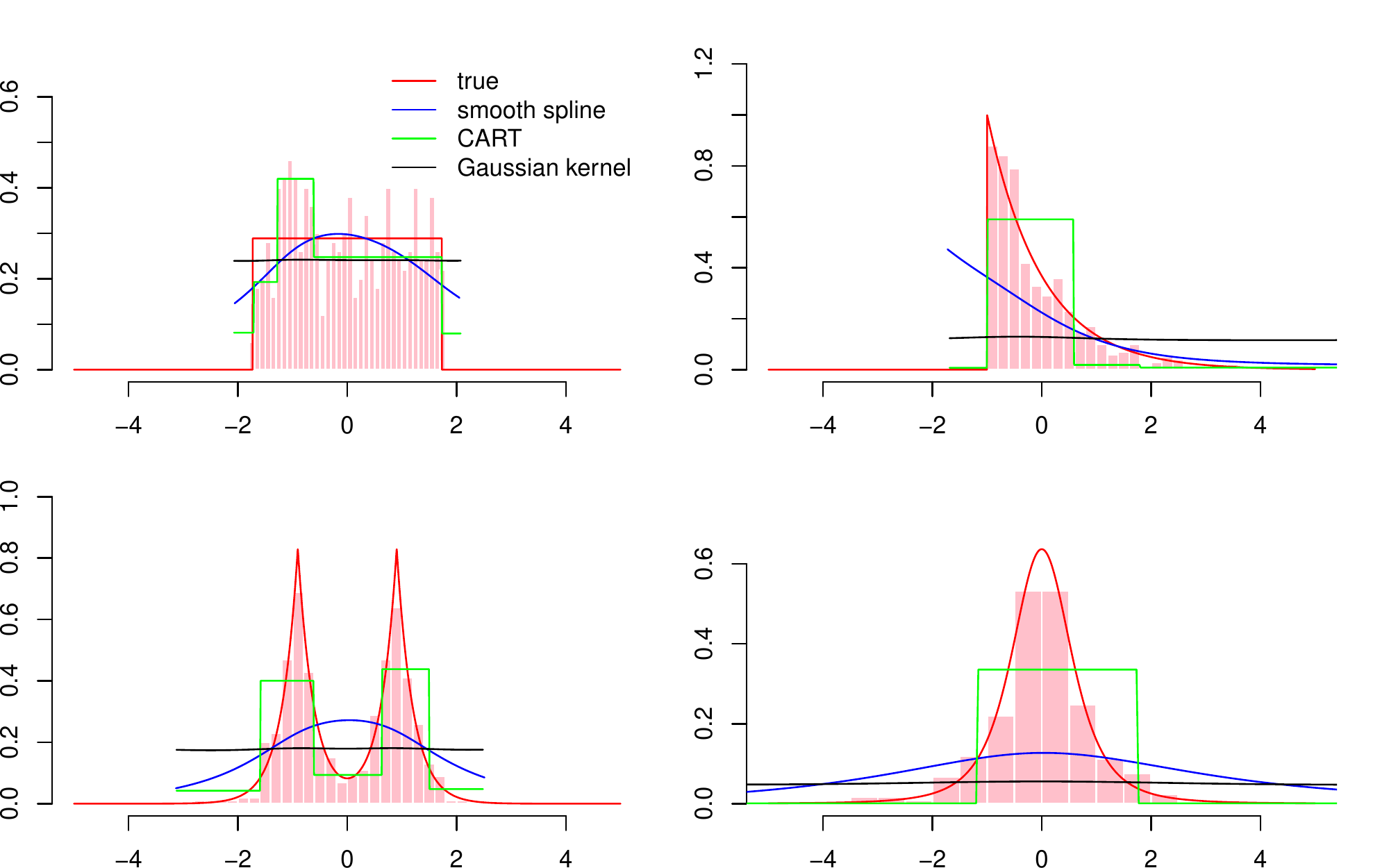}
		\label{4M1}}
	\hfil
	\subfloat[Boosting iteration $M = 200$]{\includegraphics[width=3.5in]{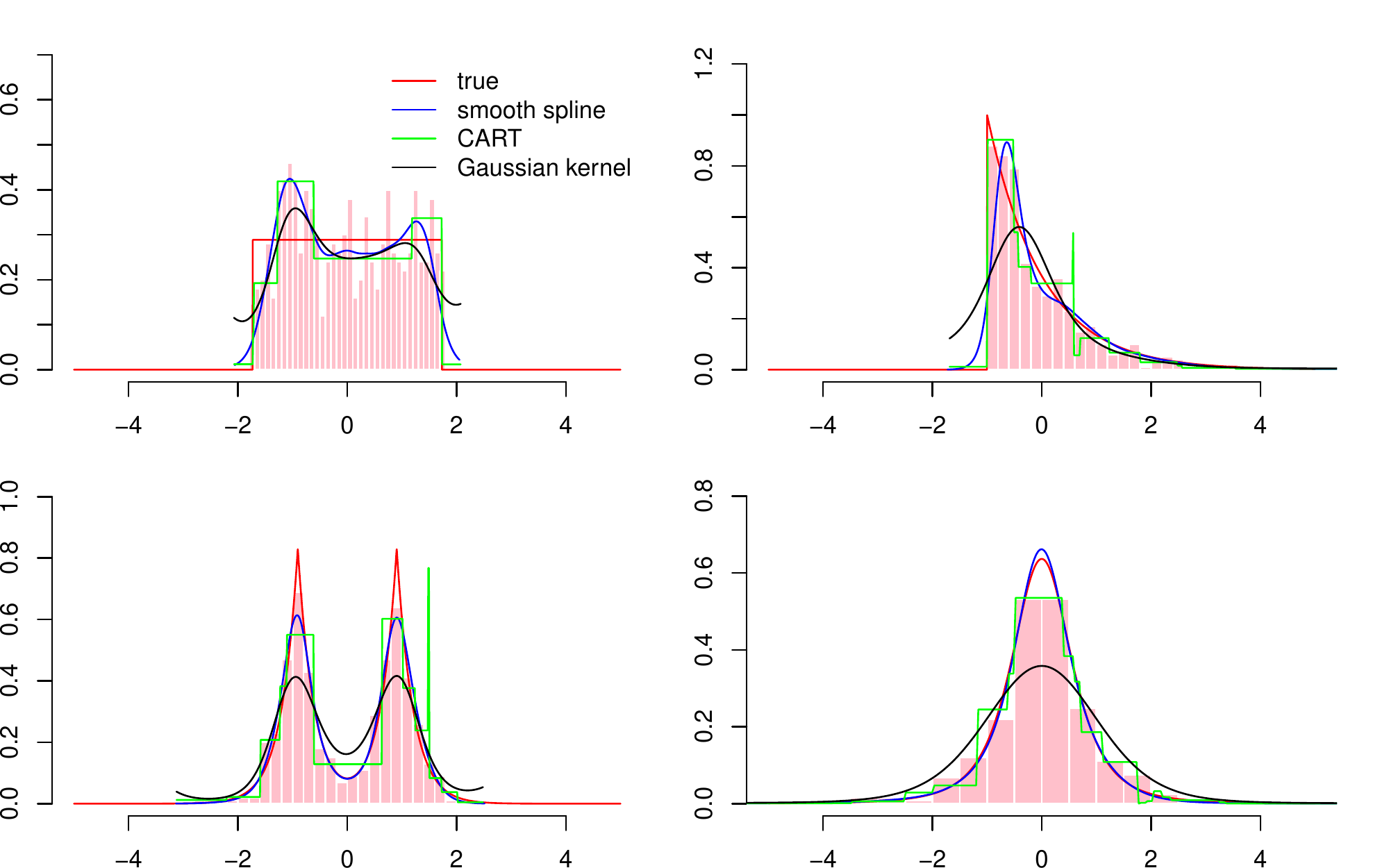}
		\label{4M200}}
	\caption{Application of boosting to the density estimation with different boosting iterations. Smooth spline (blue), CART (green), and Gaussian kernel (black) work as
		weak learners to estimate the true distributions (red), the histograms (pink) are presented for comparing. (Top left) uniform distribution; (Top right) exponential distribution; (Bottom left) mixture of two double exponential distributions; (Bottom right) student distribution.}
	\label{4M}
\end{figure*}

 In the first simulation, we apply boosting NPMLE to density estimation of different distributions, ranging from discontinuous to continuous, the sample size $N$ is $500$. As can be seen in Fig. \ref{4M1}, when $M=1$, Gaussian kernel and smooth spline do not fit well in all cases, while CART
performs well only in uniform distribution owe to its feature of piecewise constant.
 After $M=200$ boosting iterations, in Fig. \ref{4M200}, we find the estimation results of Gaussian kernel and smooth spline become closer to the ground-truth for all distributions, with performance surpassing CART. We can conclude that the performance of data fitting is significantly improved as the increase of boosting iterations $M$ for Gaussian kernel and smooth spline, and the CART is actually not appropriate to be the weak learners in boosting NPMLE.
\subsection{Robustness to the choice of M}
Although large $M$ strikingly improves the data fitting in train stage, inappropriate choice of large $M$ usually leads to overfitting in prediction for ordinary boosting methods. Another simulation is conducted to evaluate the robustness of
boosting NPMLE concerning $M$. In this simulation, we use the Gaussian kernel and smooth spline as weak learners and fix the true distribution $p(x)$ as a Gaussian Mixture Model (GMM), where the sample size is $N=500$.
\begin{equation}
\label{gmm}
p(x) =\beta\phi(x;\mu_{1},\sigma_{1}^{2}) +(1-\beta)\phi(x;\mu_{2},\sigma_{2}^{2})
\end{equation}
\begin{equation}
\label{gauss}
\phi(x;\mu,\sigma^{2})=\frac{1}{\sqrt{2\pi}\sigma}e^{-\frac{(x-\mu)^{2}}{2\sigma^{2}}}
\end{equation}
where $\beta \in [0,1],\, \mu_{1,2}=\pm2.5,\,\sigma_{1,2}^{2}=2$. We increase the $\beta$ from 0 to 1 and choose different $M$ in boosting NPMLE. The KL divergence $D_{KL}(p||\hat{p})$ is used to measure the distance between the true distribution $p(x)$ and the estimation $\hat{p}(x)$.
\begin{equation}
\label{kldivergence}
D_{KL}(p||\hat{p}) = \int p(x)log\,\frac{p(x)}{\hat{p}(x)}dx
\end{equation}
\begin{figure}[!t]
	\centering
	\subfloat[smooth spline]{\includegraphics[width=1.68in]{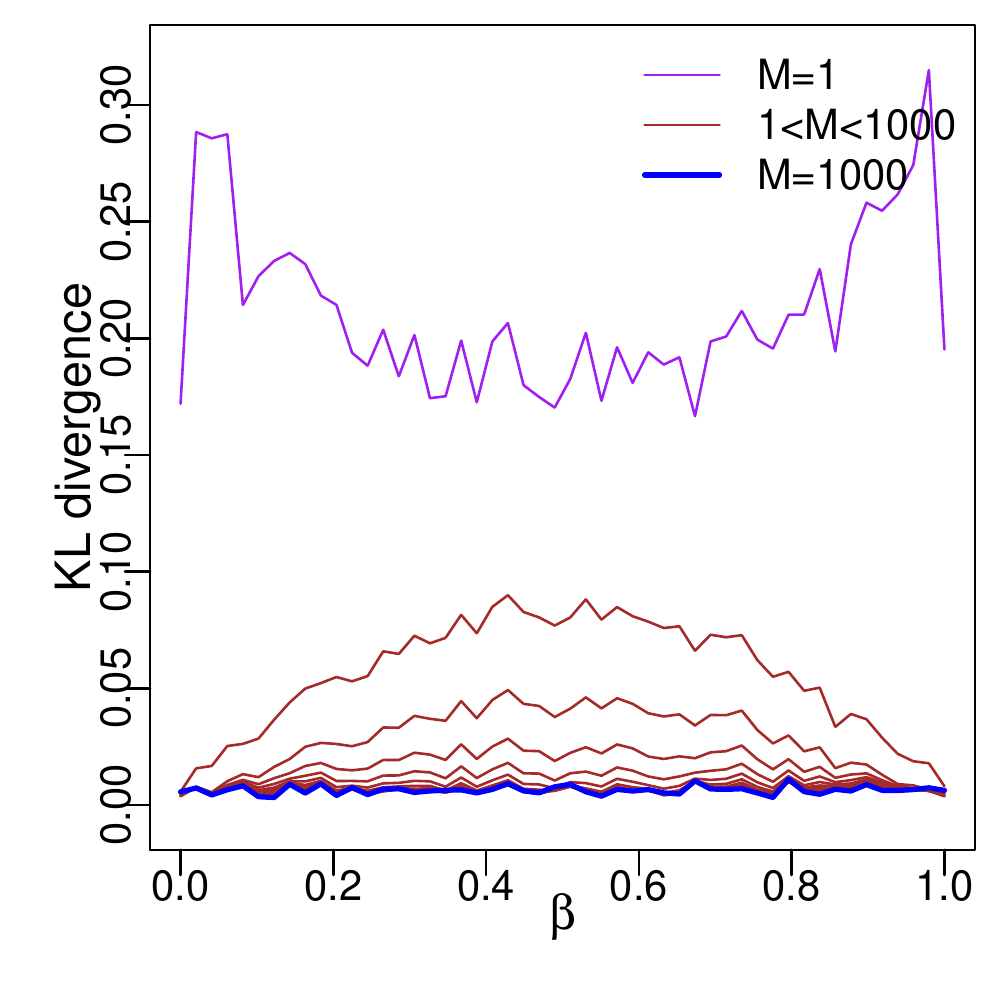}
		\label{Smooth spline}}
	\hfil
	\subfloat[Gaussian kernel]{\includegraphics[width=1.68in]{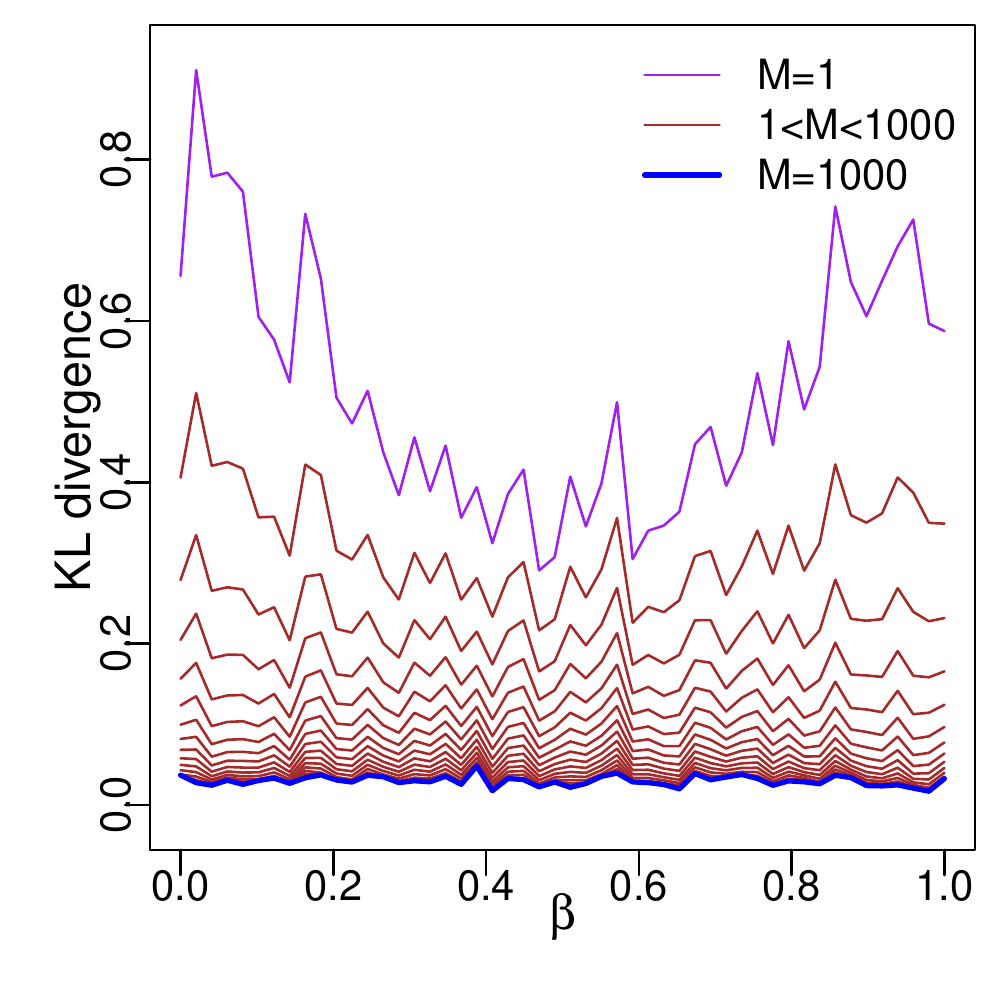}
		\label{Kernel}}
	\caption{Average KL divergence of boosting NPMLE in GMM based on 50 simulations. The ordinate is the KL divergence $D_{KL}(p||\hat{p})$, and the abscissa represents the increasing sequence concerning $\beta$. In the left figure, smooth spline works as weak learners, while Gaussian kernel works as
		weak learners in right figure. In both figures, the corresponding weak learners have different number of iterations $M$ such as $M=1$ (purple), $1<M<1000$ (brown), $M=1000$ (blue).}
	\label{changeM}
\end{figure}
In Fig. \ref{changeM}, with the increase of $M$, the KL divergences approach zero and their envelopes become surprisingly denser, which indicates that a parsimonious updating strategy is adopted by our weak learners to alleviate the risk of overfitting in the remaining iterations.

\subsection{Evaluation on pattern classification}
\begin{table}[]
    \caption{Comparisons of performance of different NPMLE methods on South African Heart Disease dataset}
	\label{pcdetails}
	\centering
	\begin{tabular}{l|ll}
		\hline
		& \multicolumn{2}{l}{Misclassification Rate(\%)} \\ \cline{2-3} 
		NPMLE method           & Training set           & Testing set           \\ \hline
		log-concavity                & 31.38($\pm$1.59)            & 32.67($\pm$3.33)           \\
		kernel                  & 30.82($\pm$1.55)            & 32.63($\pm$3.37)           \\
		penalized spline                  & 30.91($\pm$1.60)            & 33.89($\pm$3.55)           \\
		boosting NPMLE(smooth spline) & \textbf{30.99($\pm$1.70)}   & \textbf{32.58($\pm$3.63)}  \\
		boosting NPMLE(Gaussian kernel) & \textbf{30.98($\pm$1.74)}   & \textbf{32.38($\pm$3.77)}  \\\hline
	\end{tabular}
\end{table}
\begin{figure}[!t]
	\centering
	\subfloat[$chd=0$]{\includegraphics[width=1.68in]{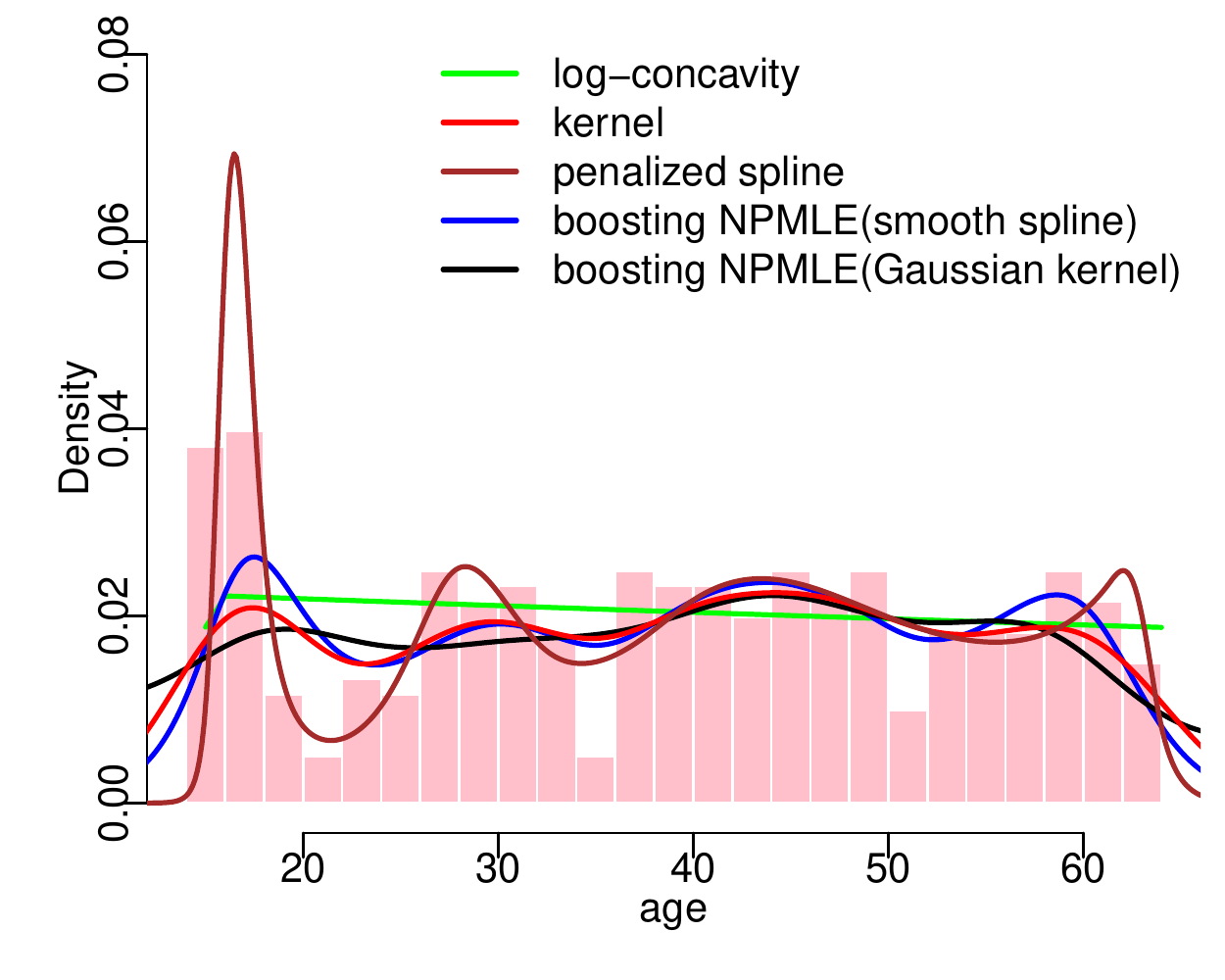}
		\label{chd}}
	\hfil
	\subfloat[$chd=1$]{\includegraphics[width=1.68in]{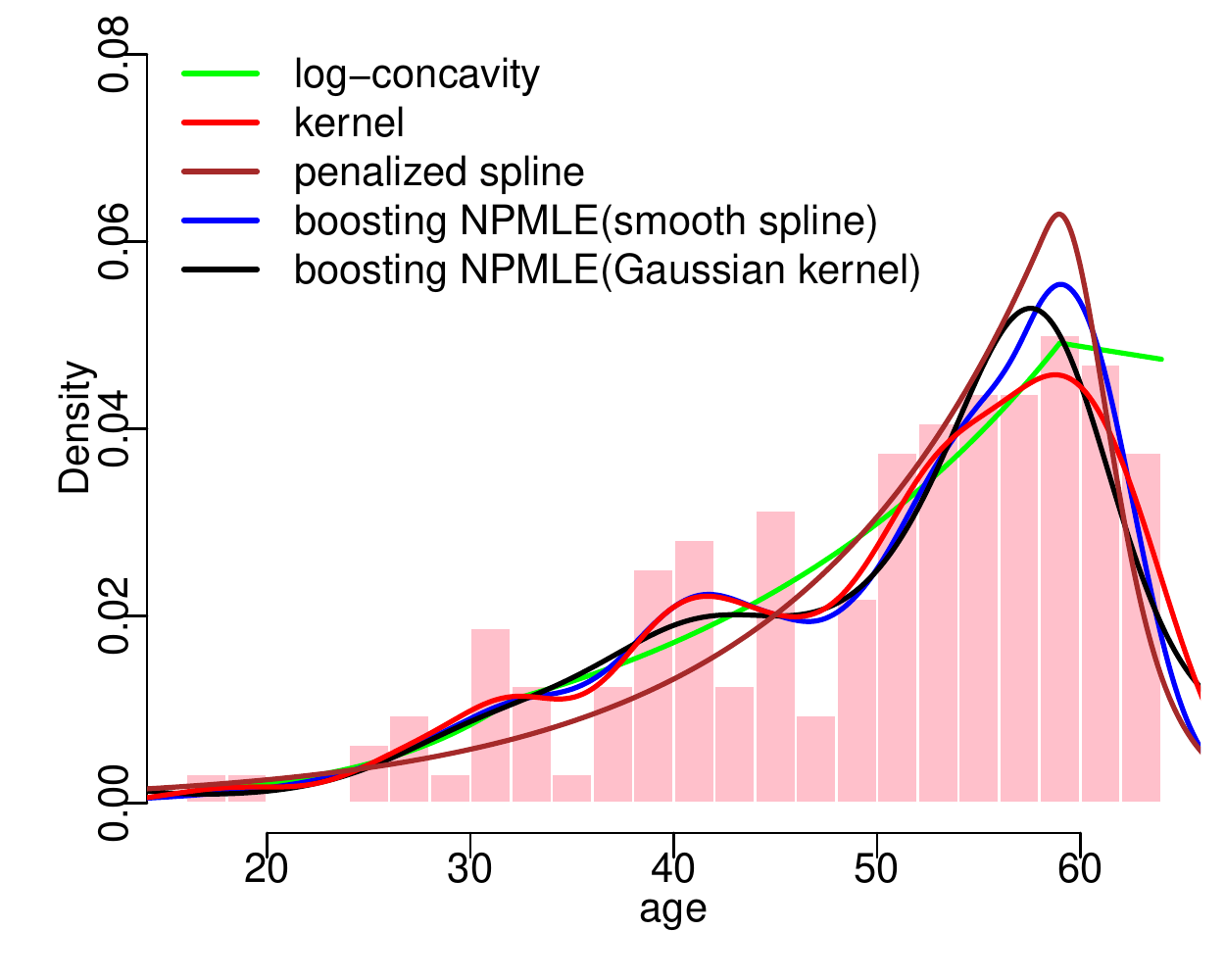}
		\label{nochd}}
	\caption{Estimated conditional densities $p(age|chd)$ to the South African Heart Disease dataset by boosting NPMLE. Histograms of $age$ for the binary response $chd$ separately.}
	\label{heart}
\end{figure}
\par We evaluate our algorithm on the South African Heart Disease dataset\cite{SAdata} for pattern classification, which contains 462 patterns ($70\%$ for the training set and $30\%$ for the testing set). We use only the quantitative input feature $age$ (age at onset) to predict the binary response $chd$ (coronary heart disease). The conditional probabilities $p(age|chd)$ are estimated by the proposed algorithms (smooth spline and Gaussian kernel) in Fig. \ref{heart}. We use bayesian classifiers to compare boosting NPMLE with other NPMLE methods including log-concavity\cite{logcondens}, kernel\cite{ks}, and penalized spline\cite{logspline} (both default parameters). Thanks to the robustness of boosting NPMLE, boosting iterations can be selected extremely large ($M=2000$). The average misclassification rate on $100$ random splits is recorded in Table \ref{pcdetails}. Our algorithm is consistent with other NPMLE methods in this task.

\section{Conclusion}
\par In this paper, a novel selection algorithm based on boosting has been proposed to solve NPMLE. We derive the boosting NPMLE by second-order approximation to log-likelihood. Different from ordinary boosting in supervised learning, our algorithm adjusts both the weight and response during the sequential routine. Several weak learners are chosen to comply with the smoothing assumptions required in NPMLE.  
Simulations and classification experiment validate the effectiveness of the proposed algorithm.

% Please add the following required packages to your document preamble:
% \usepackage{multirow}

% if have a single appendix:
%\appendix[Proof of the Zonklar Equations]
% or
%\appendix  % for no appendix heading
% do not use \section anymore after \appendix, only \section*
% is possibly needed

% use appendices with more than one appendix
% then use \section to start each appendix
% you must declare a \section before using any
% \subsection or using \label (\appendices by itself
% starts a section numbered zero.)
%

% use section* for acknowledgment
%\section*{Acknowledgment}

%The authors would like to thank...

% Can use something like this to put references on a page
% by themselves when using endfloat and the captionsoff option.
\ifCLASSOPTIONcaptionsoff
  \newpage
\fi

\bibliographystyle{IEEEtran}
\bibliography{ref}

% Generated by IEEEtran.bst, version: 1.14 (2015/08/26)
\begin{thebibliography}{10}
\providecommand{\url}[1]{#1}
\csname url@samestyle\endcsname
\providecommand{\newblock}{\relax}
\providecommand{\bibinfo}[2]{#2}
\providecommand{\BIBentrySTDinterwordspacing}{\spaceskip=0pt\relax}
\providecommand{\BIBentryALTinterwordstretchfactor}{4}
\providecommand{\BIBentryALTinterwordspacing}{\spaceskip=\fontdimen2\font plus
\BIBentryALTinterwordstretchfactor\fontdimen3\font minus
  \fontdimen4\font\relax}
\providecommand{\BIBforeignlanguage}[2]{{%
\expandafter\ifx\csname l@#1\endcsname\relax
\typeout{** WARNING: IEEEtran.bst: No hyphenation pattern has been}%
\typeout{** loaded for the language `#1'. Using the pattern for}%
\typeout{** the default language instead.}%
\else
\language=\csname l@#1\endcsname
\fi
#2}}
\providecommand{\BIBdecl}{\relax}
\BIBdecl

\bibitem{jkjw1956}
J.~Kiefer and J.~Wolfowitz, ``Consistency of the maximum likelihood estimator
  in the presence of infinitely many incidental parameters,'' \emph{The Annals
  of Mathematical Statistics}, vol.~27, no.~4, pp. 887--906, 1956.

\bibitem{steven1980}
S.~J. Bean and C.~P. Tsokos, ``Developments in nonparametric density
  estimation,'' \emph{International Statistical Review / Revue Internationale
  de Statistique}, vol.~48, no.~3, pp. 267--287, 1980.

\bibitem{alan1991}
A.~J. Izenman, ``Review papers: Recent developments in nonparametric density
  estimation,'' \emph{Journal of the American Statistical Association},
  vol.~86, no. 413, pp. 205--224, 1991.

\bibitem{eman1962}
\BIBentryALTinterwordspacing
E.~Parzen, ``On estimation of a probability density function and mode,''
  \emph{The Annals of Mathematical Statistics}, vol.~33, no.~3, pp. 1065--1076,
  1962. [Online]. Available: \url{http://www.jstor.org/stable/2237880}
\BIBentrySTDinterwordspacing

\bibitem{nada1965}
E.~A. Nadaraya, ``On non-parametric estimates of density functions and
  regression curves,'' \emph{Theory of Probability and its Applications},
  vol.~10, no.~1, pp. 186--5, 1965.

\bibitem{jeff1993}
J.~Racine, ``An efficient cross-validation algorithm for window width selection
  for nonparametric kernel regression,'' \emph{Communications in Statistics -
  Simulation and Computation}, vol.~22, no.~4, pp. 1107--1114, 1993.

\bibitem{Loader1999}
\emph{Local Regression Methods}.\hskip 1em plus 0.5em minus 0.4em\relax New
  York, NY: Springer New York, 1999.

\bibitem{matia2020}
M.~D. Cattaneo, M.~Jansson, and X.~Ma, ``Simple local polynomial density
  estimators,'' \emph{Journal of the American Statistical Association}, vol.
  115, no. 531, pp. 1449--1455, 2020.

\bibitem{nan1978}
N.~Laird, ``Nonparametric maximum likelihood estimation of a mixing
  distribution,'' \emph{Journal of the American Statistical Association},
  vol.~73, no. 364, pp. 805--811, 1978.

\bibitem{redner1984}
R.~A. Redner and H.~F. Walker, ``Mixture densities, maximum likelihood and the
  em algorithm,'' \emph{SIAM Review}, vol.~26, no.~2, pp. 195--239, 1984.

\bibitem{mich1995}
M.~D. Escobar and M.~West, ``Bayesian density estimation and inference using
  mixtures,'' \emph{Journal of the American Statistical Association}, vol.~90,
  no. 430, pp. 577--588, 1995.

\bibitem{pri2018}
P.~R. Hahn, R.~Martin, and S.~G. Walker, ``On recursive bayesian predictive
  distributions,'' \emph{Journal of the American Statistical Association}, vol.
  113, no. 523, pp. 1085--1093, 2018.

\bibitem{samworth2018}
R.~J. Samworth, ``Recent progress in log-concave density estimation,''
  \emph{Statist. Sci.}, no.~4, pp. 493--509, 11.

\bibitem{minyen1999}
M.-Y. Cheng, T.~Gasser, and P.~Hall, ``Nonparametric density estimation under
  unimodality and monotonicity constraints,'' \emph{Journal of Computational
  and Graphical Statistics}, vol.~8, no.~1, pp. 1--21, 1999.

\bibitem{goodd1971}
I.~J. GOODD and R.~A. GASKINS, ``{Nonparametric roughness penalties for
  probability densities},'' \emph{Biometrika}, vol.~58, no.~2, pp. 255--277, 08
  1971.

\bibitem{bws1982}
B.~W. Silverman, ``On the estimation of a probability density function by the
  maximum penalized likelihood method,'' \emph{The Annals of Statistics},
  vol.~10, no.~3, pp. 795--810, 1982.

\bibitem{bunea2007}
F.~Bunea, A.~Tsybakov, and M.~Wegkamp, ``Sparse density estimation with l1
  penalties,'' 06 2007, pp. 530--543.

\bibitem{koenker2006}
R.~Koenker and I.~Mizera, ``Density estimation by total variation
  regularization,'' 01 2006.

\bibitem{demontricher1975}
G.~F. de~Montricher, R.~A. Tapia, and J.~R. Thompson, ``Nonparametric maximum
  likelihood estimation of probability densities by penalty function methods,''
  \emph{Ann. Statist.}, vol.~3, no.~6, pp. 1329--1348, 11 1975.

\bibitem{bookSpline}
C.~de~Boor, \emph{A Practical Guide to Spline}, 01 1978, vol. Volume 27.

\bibitem{Edward1983}
E.~J. Wegman and I.~W. Wright, ``Splines in statistics,'' \emph{Journal of the
  American Statistical Association}, vol.~78, no. 382, pp. 351--365, 1983.

\bibitem{Gu03}
C.~Gu and J.~Wang, ``Penalized likelihood density estimation: Direct
  cross-validation and scalable approximation,'' \emph{Statistica Sinica},
  vol.~13, pp. 811--826, 2003.

\bibitem{boostingref}
P.~Bühlmann and T.~Hothorn, ``Boosting algorithms: Regularization, prediction
  and model fitting,'' \emph{Statist. Sci.}, vol.~22, no.~4, pp. 477--505, 11
  2007.

\bibitem{FREUND1997119}
Y.~Freund and R.~E. Schapire, ``A decision-theoretic generalization of on-line
  learning and an application to boosting,'' \emph{Journal of Computer and
  System Sciences}, vol.~55, no.~1, pp. 119 -- 139, 1997.

\bibitem{friedman2001}
J.~H. Friedman, ``Greedy function approximation: A gradient boosting machine.''
  \emph{Ann. Statist.}, vol.~29, no.~5, pp. 1189--1232, 10 2001.

\bibitem{glmnet2011}
\BIBentryALTinterwordspacing
N.~Simon, J.~Friedman, T.~Hastie, and R.~Tibshirani, ``Regularization paths for
  cox's proportional hazards model via coordinate descent,'' \emph{Journal of
  Statistical Software}, vol.~39, no.~5, pp. 1--13, 2011. [Online]. Available:
  \url{http://www.jstatsoft.org/v39/i05/}
\BIBentrySTDinterwordspacing

\bibitem{SAdata}
\BIBentryALTinterwordspacing
T.~Hastie, R.~Tibshirani, and J.~Friedman, ``The elements of statistical
  learning -- data mining, inference, and prediction.''\hskip 1em plus 0.5em
  minus 0.4em\relax Springer New York, 2009, ch.~6, pp. 214--215. [Online].
  Available:
  \url{https://web.stanford.edu/~hastie/ElemStatLearn/datasets/SAheart.data}
\BIBentrySTDinterwordspacing

\bibitem{logcondens}
\BIBentryALTinterwordspacing
L.~D\"umbgen and K.~Rufibach, ``{logcondens}: Computations related to
  univariate log-concave density estimation,'' \emph{Journal of Statistical
  Software}, vol.~39, no.~6, pp. 1--28, 2011. [Online]. Available:
  \url{http://www.jstatsoft.org/v39/i06/}
\BIBentrySTDinterwordspacing

\bibitem{ks}
\BIBentryALTinterwordspacing
T.~Duong, \emph{ks: Kernel Smoothing}, 2020, r package version 1.11.7.
  [Online]. Available: \url{https://CRAN.R-project.org/package=ks}
\BIBentrySTDinterwordspacing

\bibitem{logspline}
\BIBentryALTinterwordspacing
C.~Kooperberg, \emph{logspline: Routines for Logspline Density Estimation},
  2020, r package version 2.1.16. [Online]. Available:
  \url{https://CRAN.R-project.org/package=logspline}
\BIBentrySTDinterwordspacing

\end{thebibliography}

% that's all folks
\end{document}